\documentclass[runningheads,a4paper]{llncs}

\usepackage{amssymb}
\usepackage{graphicx}
\usepackage{amsmath}

\usepackage{url}
\urldef{\mailsa}\path|{zhoujp877, suwh, ymh}@nenu.edu.cn|
\newcommand{\keywords}[1]{\par\addvspace\baselineskip
\noindent\keywordname\enspace\ignorespaces#1}

\begin{document}

\mainmatter  

\title{Approximate Counting CSP \\Solutions Using Partition Function}

\titlerunning{Approximate Counting CSP Solutions Using Partition Function}

%
%
\author{Zhou Junping%
\and Su Weihua%
\and Yin Minghao%
}
\authorrunning{Zhou, Su, and Yin}

\institute{College of Computer, Northeast Normal University,\\
130117, Changchun, P. R. China\\
\mailsa\\
}

%
%

\toctitle{Lecture Notes in Computer Science}
\tocauthor{Authors' Instructions}
\maketitle

\begin{abstract}
We propose a new approximate method for counting the number of the
solutions for constraint satisfaction problem (CSP). The method
derives from the partition function based on introducing the free
energy and capturing the relationship of probabilities of variables
and constraints, which requires the marginal probabilities. It
firstly obtains the marginal probabilities using the belief
propagation, and then computes the number of solutions according to
the partition function. This allows us to directly plug the marginal
probabilities into the partition function and efficiently count the
number of solutions for CSP. The experimental results show that our
method can solve both random problems and structural problems
efficiently. \keywords{Partition function; \#CSP; belief
propagation; marginal probability.}
\end{abstract}

\section{Introduction}

Counting the number of solutions for constraint satisfaction
problem, denoted by \#CSP, is a very important problem in Artificial
Intelligence (AI). In theory, \#CSP is a \#P-complete problem even
if the constraints are binary, which has played a key role in
complexity theory. In practice, effective counters have opened up a
range of applications, involving various probabilistic inferences,
approximate reasoning, diagnosis, and belief revision.

In recent years, many attentions have been focused on counting a
specific case of \#CSP, called \#SAT. By counting components,
Bayardo and Pehoushek presented an exact counter for SAT, called
Relsat [1]. By combining component caching with clause learning
together, Sang et al. created an exact counter cachet [2]. Based on
converting the given CNF formula into d-DNNF form, which makes the
counting easily, Darwiche introduced an exact counter c2d [3]. By
introducing an entirely new approach of coding components, Thurley
addressed an exact counter sharpSAT [4]. By using more reasoning,
Davies and Bacchus addressed an exact counter \#2clseq [5]. Besides
the emerging exact \#SAT solvers, Wei and Selman presented an
approximate counter ApproxCount for SAT by using Markov Chain Monte
Carlo (MCMC) sampling [6]. Building upon ApproxCount, Gomes et al.
used sampling with a modified strategy and proposed an approximate
counter SampleCount [7]. Relying on the properties of random XOR
constraints, an approximate counter MBound was introduced in [8].
Using sampling of the backtrack-free search space of systematic SAT
solver, SampleMinisat was addressed in [9]. Building on the
framework of SampleCount, Kroc et al. exploited the belief
propagation method and presented an approximate counter BPCount
[10]. By performing multiple runs of the MiniSat SAT solver, Kroc et
al. introduced an approximate counter, called MiniCount [10].

Recently, more efforts have been made on the general \#CSP problems.
For example, Angelsmark et al. presented upper bounds of the \#CSP
problems [11]. Bulatov and Dalmau discussed the dichotomy theorem
for the counting CSP [12]. Pesant exploited the structure of the CSP
models and addressed an algorithm for solving \#CSP [13]. Dyer et
al. considered the trichotomy theorem for the complexity of
approximately counting the number of satisfying assignments of a
Boolean CSP [14]. Yamakami studied the dichotomy theorem of
approximate counting for complex-weighted Boolean CSP [15]. Though
great many studies had been made on the algorithms for the \#CSP
problems, only a few of them related to the \#CSP solvers. Gomes et
al. proposed a new generic counting technique for CSPs building upon
on the XOR constraint [16]. By adapting backtracking with
tree-decomposition, Favier et al. introduced an exact \#CSP solver,
called \#BTD [17]. In addition, by relaxing the original CSP
problems, they presented an approximate method Approx \#BTD [17].

In this paper, we propose a new type of method for solving \#CSP
problems. The method derives from the partition function based on
introducing the free energy and capturing the relationship of
probabilities of variables and constraints. When computing the
number of the solutions of a given CSP formula according to the
partition function, we require the marginal probabilities of each
variable and each constraint to plug into the partition function. In
order to obtain the marginal probabilities, we employ the belief
propagation (BP) because it can organize a computation that makes
the marginal probabilities computing tractable and eventually
returns the marginal probabilities. In addition, unlike the counter
BPCount using the belief propagation method for obtaining the
information deduced from solution samples in SampleCount, we employ
the belief propagation method for acquiring information for
partition function. This leads to two differences between BPCount
and our counter. The first one is the counter BPCount requires to
iteratively perform the belief propagation method and repeatedly
obtain the marginal probabilities of each variable on the simplified
SAT formulae; while our counter carries out the belief propagation
method only once, which spends less cost. The second one is that the
two counters obtaining the exact number of solutions depending on
different circumstances. The counter BPCount needs the corresponding
factor graphs of the simplified SAT formulae all have no cycles;
while our counter only needs the factor graph of the given CSP
formula has no cycle, which meets easily.

Our experiments reveal that our counter for CSP, called PFCount,
works quite well. We consider various hard instances, including the
random instances and the structural instances. For the random
instances, we consider the instances based on the model RB close to
the phase transition point, which has been proved the existence of
satisfiability phase transition and identified the phase transition
points exactly. With regard to the random instances, our counter
PFCount improves the efficiency tremendously especially for
instances with more variables. Moreover, PFCount presents a good
estimate to the number of solutions for instances based on model RB,
even if the instances scales are relatively large. Therefore, the
effectiveness of PFCount is much more evident especially for random
instances. For the structural instances, we focus on the counting
problem based on graph coloring. The performance of PFCount for
solving structural instances is in general comparing with the random
instances because PFCount sometimes can't converge. However, once
PFCount can converge, it can estimate the number of the solutions of
instances efficiently. As a whole, PFCount is a quite competitive
\#CSP solver.

\section{Preliminaries}

A constraint satisfaction problem (CSP) $\mathcal {P}$ is defined as
a pair $\mathcal {P} = \langle V, C \rangle$, where $V=\{x_1, x_2,
..., x_n\}$ is a set of variables and $C=\{c_1, c_2, ..., c_m\}$ is
a set of constraints defined on \textit{V}. For each variable $x_i$
in \textit{V}, the domain $D_i$ of $x_i$ is a set with $|D_i|$
values; the variable $x_i$ can be only assigned a value from $D_i$.
A constraint \textit {c}, called a \textit {k}-ary constraint,
consists of \textit {k} variables $x_{c_1}, x_{c_2}, ..., x_{c_k}$
and a relation $R \subseteq D_{c_1} \times  D_{c_2} \times ...
\times D_{c_k}$, where $c_1$, $c_2$, ..., $c_k$ are distinct. The
relation $R$ specifies all the allowed tuples of values for the
variables $x_{c_1}, x_{c_2}, ..., x_{c_k}$ which are compatible with
each other. The variable configuration of a CSP $\mathcal {P}$ is
$X=\{x_1 = d_1, x_2 = d_2, ..., x_n = d_n\}$ that assigns each
variable a value from its domain. A solution to a constraint $c_i$
is a variable configuration $X_{c_i}=\{x_{c_1} = d_{c_1}, x_{c_2} =
d_{c_2}, ..., x_{c_k} = d_{c_k}\}$ that sets values to each variable
in the constraint $c_i$ such that $X_{c_i} \in R_i$. We also say
that the variable configuration $X_{c_i}$ satisfies the constraint
$c_i$. A solution to a CSP $\mathcal {P}$ is a variable
configuration such that all the constraints in \textit {C} are
satisfied. Given a CSP $\mathcal {P}$, the decision problem is to
determine whether the CSP $\mathcal {P}$ has a solution. The
corresponding counting problem (\#CSP) is to determine how many
solutions the CSP $\mathcal {P}$ has.

A CSP $\mathcal {P}$ can be expressed as a bipartite graph called
factor graph (see Fig. 1). The factor graph has two kinds of nodes,
one is variable node (which we draw as circles) representing the
variables, and the other is function node (which we draw as squares)
representing the constraints. A function node is connected to a
variable node by an edge if and only if the variable appeares in the
constraint. In the rest of this paper, we will always index variable
nodes with letters starting with \textit {i}, and factor nodes with
letters starting with $c_i$. In addition, for every variable node
\textit {i}, we will use \textit {V}(\textit {i}) to denote the set
of function nodes which it connects to, and \textit {V}(\textit
{i})$\setminus c_i$ to denote the set \textit {V}(\textit {i})
without function node $c_i$. Similarly, for each function node
$c_i$, we will use \textit {V}($c_i$) to denote the set of variable
nodes which it connects to, and \textit {V}($c_i$)$\setminus$
\textit {i} to denote the set \textit {V}($c_i$) without variable
node \textit {i}.

\begin{figure}
\centering
\includegraphics[scale=0.5]{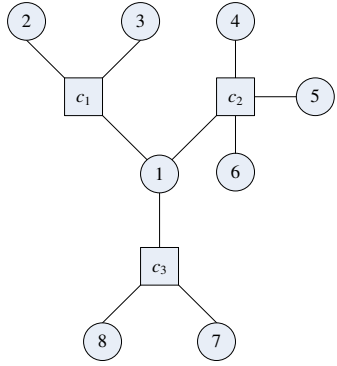}
\caption{An example of a factor graph with 8 variable nodes \textit
{i}=1, 2, ..., 8 and 3 function nodes $c_1, c_2, c_3$. The CSP
$\mathcal {P}$ is encoded as $\mathcal {P} = \langle
 \{x_1, x_2,..., x_8\}, \{c_1 = (x_1 < x_2 < x_3), c_2 = (x_1 \neq x_4
\neq x_5 \neq x_6), c_3 = (x_1 + x_7 + x_8 <10)\}\rangle$, where
\textit {V}($c_1$) = \{1, 2, 3\}, \textit {V}($c_2$) = \{1, 4, 5,
6\}, \textit {V}($c_3$) = \{1, 7, 8\}.} \label{fig.example}
\end{figure}

\section{Partition Function for Solving \#CSP}

In this section, we present a new approximate approach, called
PFCount, for counting the number of solutions for constraint
satisfaction problem. The approach derives from the partition
function based on introducing the free energy and capturing the
relationship of probabilities of variables and constraints. In the
following, we will describe the partition function in details.

\subsection{Partition Function for Counting}

In this subsection, we present a partition function for counting the
number of solutions for CSP. The partition function is an important
quantity in statistical physics, which describes the statistical
properties of a system. Most of the aggregate thermodynamic
variables of the system, such as the total energy, free energy,
entropy, and pressure, can be expressed in terms of the partition
function. To facilitate the understanding, we first describe the
notion of the partition function. Given a system of \textit {n}
particles, each of which can be in one of a discrete number of
states, i.e., {$d_1, d_2 ,..., d_n$}, and a state of the system
\textit {X} denoted by $X=\{x_1 = d_1, x_2 = d_2, ..., x_n = d_n\}$,
i.e., the \textit{i}th particle $x_i$ is in the state $d_i$, the
partition function in statistical physics is defined as

\begin{equation}
  Z(T)={1\over p(X)} e^{-E(X)/T}
\end{equation}where \textit {T} is the temperature, \textit {E}(\textit
{X}) is the energy of the state \textit {X},  and \textit
{p}(\textit {X}) is the probability of the state \textit {X}. In
this paper, we focus on the partition function that the temperature
\textit {T} is assigned to 1.

Since the partition function is also used in probability theory, in
the following we will learn the partition function from the
probability theory. Given a CSP $\mathcal {P}$ and a variable
configuration $X=\{x_1 = d_1, x_2 = d_2, ..., x_n = d_n\}$ of
$\mathcal {P}$, the partition function in probability theory is
defined in Equation (2).

\begin{equation}
  Z={1\over p(X)} \prod_{i=1}^m f_{c_i}(X)
\end{equation} where \textit {p}(\textit {X}) is the the joint probability distribution,
function $f_{c_i}(X)$ is a Boolean function range \{0, 1\}, which
evaluates to 1 if and only if the constraint $c_i$ is satisfied,
evaluates to 0 otherwise; and \textit {m} is the number of
constraints. Based on Equation (2), the joint probability
distribution \textit {p}(\textit {X}) over the \textit {n} variables
can be expressed as the follows.

\begin{equation}
  p(X)={1\over Z} \prod_{i=1}^m f_{c_i}(X)
\end{equation}Because the construction of the joint probability distribution is
uniform over all variable configurations, \textit {Z} is the number
of solutions of the given CSP $\mathcal {P}$. Therefore, \#CSP can
be solved by computing a partition function. In the following, we
will propose the derivation of the partition function.

In order to present a calculation method to compute the partition
function, we introduce the variational free energy defined by

\begin{equation}
  F(b(X))=\sum_X b(X)E(X) + \sum_X b(X)\ln b(X)
\end{equation} where \textit {E}(\textit {X}) is the energy of the
state \textit {X} and \textit {b}(\textit {X}) is a trial
probability distribution. Simplifying the Equation (4), we draw up
the following equation.

\begin{eqnarray}
  F(b(X))&=&\sum_X b(X)E(X) + \sum_X b(X)\ln b(X) \nonumber  \\
         &=&\sum_X b(X) \ln (e^{E(X)}b(X)) \nonumber  \\
         &=&\sum_X b(X) \ln{b(X) \over e^{-E(X)}}
\end{eqnarray}By setting \textit {T} to 1 in Equation (1), we can obtain:

\begin{equation}
  e^{-E(X)}=p(X)Z
\end{equation}Then we take the Equation (6) into (5) and acquire:

\begin{eqnarray}
  F(b(X))&=&\sum_X b(X) \ln{b(X) \over p(X)Z} \nonumber  \\
         &=&\sum_X b(X) (\ln{b(X) \over p(X)} - \ln Z)  \nonumber  \\
         &=&-\ln Z \sum_X b(X) + \sum_X b(X) \ln{b(X) \over p(X)}
\end{eqnarray}Since \textit {b}(\textit {X}) is a trial probability distribution,
the sum of the probability distribution should be 1, i.e. $\sum_{X}
b(X)=1$. Then the Equation (7) can be expressed as the follows.

\begin{equation}
  F(b(X))=-\ln Z + \sum_X b(X) \ln {b(X) \over p(X)}
\end{equation}By analyzing the Equation (8), we know that the second term is equal
to zero if \textit {b}(\textit {X}) is equal to \textit {p}(\textit
{X}). So when \textit {b}(\textit {X}) is equal to \textit
{p}(\textit {X}), the partition function can be written as

\begin{equation}
  Z=\exp (-F(p(X)))
\end{equation} Then by taking the Equation (4) into the above
equation, we obtain

\begin{equation}
  Z=\exp (-\sum_X p(X)E(X) - \sum_X p(X)\ln p(X))
\end{equation}

For a factor graph with no cycles, \textit {p}(\textit {X}) can be
easily expressed in terms of the  marginal probabilities of
variables and constraints as the follows.

\begin{equation}
  p(X)=\prod_{i=1}^m p_{c_i}(X_{c_i}) \prod_{j=1}^n p_j (d_j)^{1- \sigma_j}
\end{equation}where $\sigma_j$ is the number of times that the variable $x_j$ occurs in the
constraints, \textit {m} and \textit {n} are the number of
constraints and variables respectively, $p_{c_i}(X_{c_i})$ and $p_j
(d_j)$ are the marginal probabilities of constraints and variables
respectively.

In addition, by analyzing the two partition functions presented in
equations (1) and (2), we can see that \textit {p}(\textit {X}) and
\textit {Z} are equal when \textit {T} is set to 1. Thus, we can
obtain the following equation from Equation (1) and Equation (2) on
account of the equivalents \textit {Z} and \textit {p}(\textit {X}).

\begin{equation}
  E(X)=-\ln \prod_{i=1}^m f_{c_i}(X)=-\sum_{i=1}^m \ln f_{c_i}(X)
\end{equation}Then the partition function can be expressed as the follows by
plugging the Equation (11) and Equation (12) into Equation (10).

\begin{equation}
\begin{aligned}
 Z =\exp &(-\sum_X(\prod_{i=1}^m p_{c_i}(X_{c_i})(\prod_{j=1}^n p_j (d_j)^{1-
  \sigma_j})(-\sum_{i=1}^m\ln f_{c_i}(X_{c_i})))\\
  &-\sum_X (\prod_{i=1}^m p_{c_i}(X_{c_i}) \prod_{j=1}^n p_j (d_j)^{1- \sigma_j})\ln (\prod_{i=1}^m p_{c_i}(X_{c_i}) \prod_{j=1}^n p_j (d_j)^{1- \sigma_j}))
\end{aligned}
\end{equation}In Equation (13), when the variable configuration \textit {X} is a
solution to a CSP $\mathcal {P}$, the function $f_{c_i}(X_{c_i})$ is
assigned 1, which means that the term $-\sum_{i=1}^m\ln
f_{c_i}(X_{c_i})$ evaluates to 0. On the other hand, when the
variable configuration \textit {X} is not a solution to a CSP
$\mathcal {P}$, the term $\prod_{i=1}^m p_{c_i}(X_{c_i})$ evaluates
to 0. Therefore, whether or not the variable configuration \textit
{X} is a solution to a CSP $\mathcal {P}$, the first term in the
exponential function must evaluate to 0. Then Equation (13) can be
expressed as the follows.

\begin{equation}
\begin{aligned}
 Z=&\exp(-\sum_X (\prod_{i=1}^m p_{c_i}(X_{c_i}) \prod_{j=1}^n p_j (d_j)^{1- \sigma_j})\ln (\prod_{i=1}^m p_{c_i}(X_{c_i}) \prod_{j=1}^n p_j (d_j)^{1-
 \sigma_j}))\\
  =&\exp(-\sum_{i=1}^m \sum_{X_{c_i}}p_{c_i}(X_{c_i}) \ln p_{c_i}(X_{c_i})-\sum_{j=1}^n (1-\sigma_j)\sum_{d_j}p_j(d_j)\ln p_j(d_j))
\end{aligned}
\end{equation}

From the above equation, we can learn that the number of solutions
of a given CSP $\mathcal {P}$ can be calculated according to the
partition function if the marginal probability of variable
$p_j(d_j)$ and the marginal probability of constraint
$p_{c_i}(X_{c_i})$ can be obtained. In the following, we will
present an approach to compute the marginal probabilities.

\subsection{Marginal Probabilities Estimate Using BP}

In this subsection, we address a method BP to calculate the marginal
probabilities. The belief propagation, BP for short, is a message
passing procedure, which is a method for computing marginal
probabilities [18]. The BP procedure obtains exact marginal
probabilities if the factor graph of the given CSP $\mathcal {P}$
has no cycles, and it can still empirically provide good approximate
results even when the corresponding factor graph does have cycles.

\begin{figure} \centering
\includegraphics[scale=1.0]{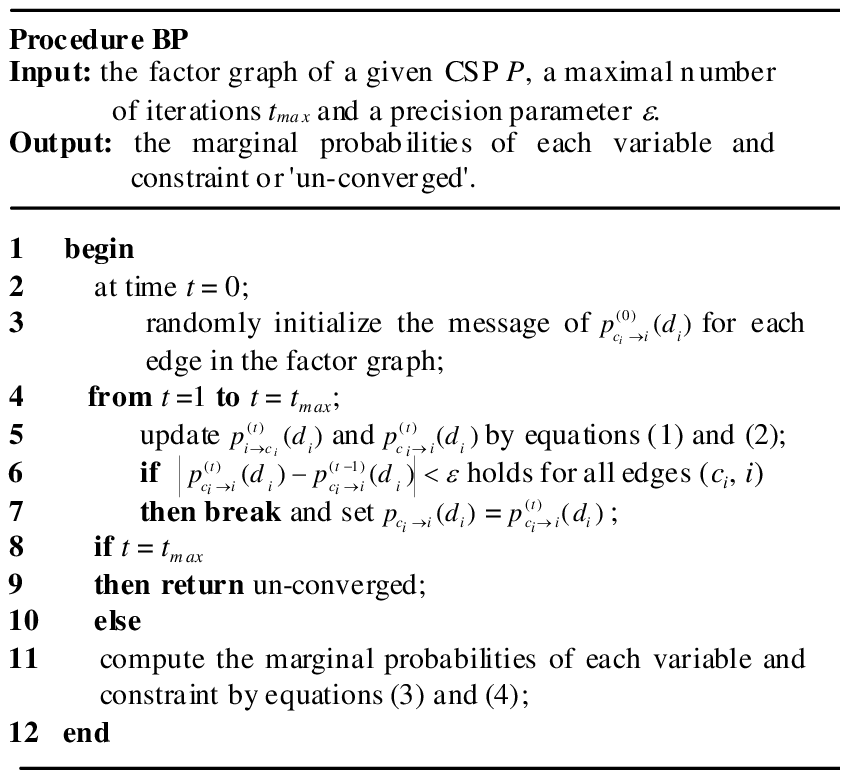}
\caption{The BP procedure.} \label{fig.example}
\end{figure}

To describe the BP procedure, we first introduce messages between
function nodes and their neighboring variable nodes and vice versa.
The message $p_{c_i\rightarrow i}(d_i)$ passed from a function node
$c_i$ to one of its neighboring variable nodes \textit {i} can be
interpreted as the probability of constraint $c_i$ being satisfied
if the variable $x_i$ takes the value $d_i\in D_i$; while the
message $p_{i\rightarrow c_i}(d_i)$ passed from a variable node
\textit {i} to one of its neighboring function nodes $c_i$ can be
interpreted as the probability that the variable $x_i$ takes the
special value $d_i\in D_i$ in the absence of constraint $c_i$. Next
we concentrate on presenting the details of the BP procedure (see
Fig. 2). At first, the message $p_{c_i\rightarrow i}(d_i)\in [0, 1]$
is initialized for every edge ($c_i$, \textit {i}) and every value
$d_i\in D_i$. Then the messages are updated with the following
equations.

\begin{equation}
  p_{i\rightarrow c_i}^{(t)}(d_i)=C_{i\rightarrow c_i} \prod_{c_j \in V(i)\setminus c_i} p_{c_j\rightarrow i}^{(t-1)}(d_i)
\end{equation}

\begin{equation}
  p_{c_i\rightarrow i}^{(t)}(d_i)=\sum_{d_j \in D_j, j \in V(c_i)\setminus i} f_{c_i}(X_{c_i}) \prod_{j \in V(c_i)\setminus i} p_{j\rightarrow c_i}^{(t-1)}(d_i)
\end{equation} where $C_{i\rightarrow c_i}$ is a normalization constant ensuring that
$p_{i\rightarrow c_i}(d_i)$ is a probability, and $f_{c_i}(X_{c_i})$
is a characteristic function taking the value 1 if the variable
configuration $X_{c_i}$ satisfies the constraint $c_i$, taking the
value 0 otherwise. The BP procedure  runs the equations (15) and
(16) iteratively until the message $p_{c_i\rightarrow i}(d_i)$
converges for every edge ($c_i$, \textit {i}) and every value
$d_i\in D_i$. When they have converged, we can then calculate the
marginal probabilities of each variable and each constraint in the
following equations.

\begin{equation}
  p_i(d_i)=C' \prod_{c_i \in V(i)} p_{c_i\rightarrow i}(d_i)
\end{equation}

\begin{equation}
  p_{c_i}(X_{c_i})=C\texttt{"} \prod_{i \in V(c_i)} \prod_{c_j \in V(i)\setminus c_i} p_{c_j\rightarrow i}(d_i)
\end{equation}where \textit {C}\texttt' and \textit {C\texttt{"}} are normalization constants ensuring that $p_i(d_i)$ and $p_{c_i}(X_{c_i})$ are
probabilities, and  $f_{c_i}(X_{c_i})$ is a characteristic function
taking the value 1 if the variable configuration $X_{c_i}$ satisfies
the constraint $c_i$, taking the value 0 otherwise.

As a whole, the BP procedure organizes a computation that makes the
marginal probabilities computing tractable and eventually returns
the marginal probabilities of each variable and each constraint
which can be used in the partition function. As we know, the BP
procedure can present exact marginal probabilities if the factor
graph of the given CSP $\mathcal {P}$ has no cycle. And from the
whole derivation of the partition function, we understand that all
equations address exact results if the factor graph is a tree. Thus,
we obtain the following theorem.

\subsubsection {Theorem 1}
The method PFCount provides an
exact number of solutions for a CSP $\mathcal {P}$ if the factor
graph of the given CSP $\mathcal {P}$ has no cycle.

The above theorem illustrates that PFCount can present an exact
number of solutions if the corresponding factor graph of the  given
CSP $\mathcal {P}$ has no cycle. In addition, even when the factor
graph does have cycles, our method still empirically presents good
approximate number of solutions for CSP.

\section{Experimental Results}

In this section, we perform two experiments on a cluster of 2.4 GHz
Intel Xeon machines with 2 GB memory running Linux CentOS 5.4. The
purpose of the first experiment is to demonstrate the performance of
our method on random instances; the second experiment is to compare
our method with two other methods on structural instances. Our \#CSP
solver is implemented in C++, which we also call PFCount. For each
instance, the run-time is in seconds and the timeout limit is 7200s.

\subsection{Evaluation on the Random Instances}

In this subsection, we conduct experiments on CSP benchmarks of
model RB, which can provide a framework for generating
asymptotically hard instances so as to give a challenge for
experimental evaluation of the \#CSP solvers [19]. The benchmarks of
model RB is determined by parameters (\textit {k}, \textit {n},
$\alpha$, \textit {r}, \textit {p}), where \textit {k} denotes the
arity of each constraint; \textit {n} denotes the number of
variables; $\alpha$ determines the domain size $d=n^\alpha$ of each
variable; \textit {r} determines the number $m=rn\ln n$ of
constraints; \textit {p} determines the number $t = pd^k$ of
disallowed tuples of each relation.

\subsubsection{Comparing PFCount with State-of-the-Art
Counters}

\begin{table}
\centering
\includegraphics[scale=0.80]{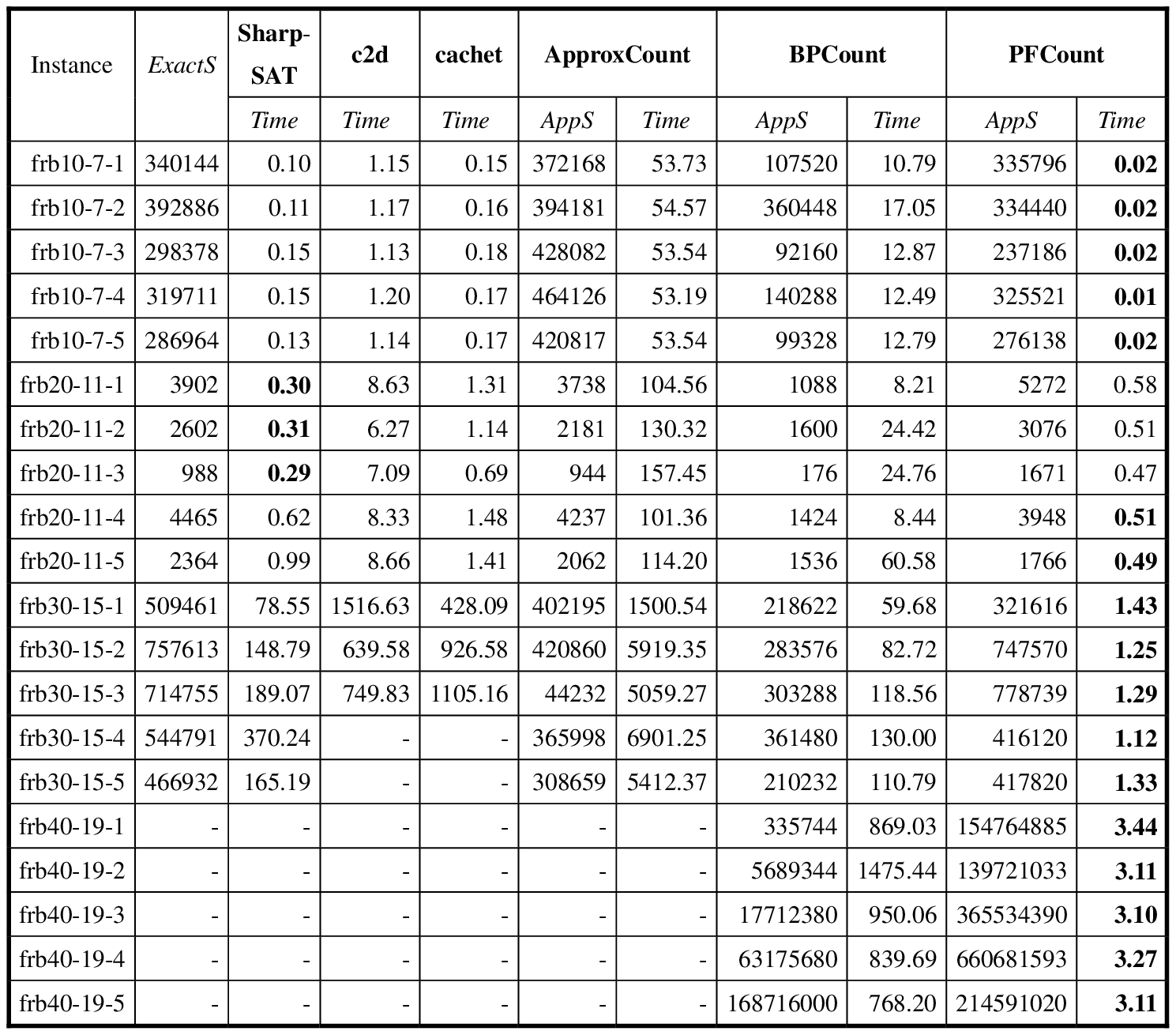}
\caption{Comparison of counters on CSP benchmarks of model RB (-
represents out of time and times are in seconds). }
\label{fig.example}
\end{table}

Table 1 illustrates the comparison of PFCount with state-of-the-art
exact \#SAT solvers sharpSAT, c2d, cachet, and approximate solvers
ApproxCount, BPCount on CSP benchmarks of model RB close to the
phase transition point. In the experiment, we choose random
instances $\mathcal {P} \in$ (2, \textit {n}, 0.8, 3, \textit {p})
with \textit {n} $\in$ \{10, 20, 30, 40\}. Moreover, since the
theoretical phase transition point $p\approx$ 0.23 for \textit {k} =
2, $\alpha$ = 0.8, \textit {r} = 3, \textit {n} $\in$ \{10, 20, 30,
40\}, we set \textit {p} = 0.20. In Table 1, the instance frb\textit
{a}-\textit {b}-\textit {c} represents the instance containing
\textit {a} variables, owning a domain with \textit {b} values for
each variable, and indexing \textit {c}; \textit {ExactS} represents
the exact number of solutions of each instance; \textit {AppS}
represents the approximate number of solutions of each instance.
Note that the exact number of solutions of each instance is obtained
by the exact \#SAT solvers and the CSP instances solved by these
\#SAT solvers are translated into SAT instances using the direct
encoding method. The results reported in Table 1 suggest that the
effectiveness of PFCount is much more evident especially for larger
instances. For example, the efficiency of solving instances with 30
variables has been raised at least 74 times (instance frb30-15-1).
And the instances with 40 variables can be solved by PFCount in a
few seconds. Furthermore, PFCount presents a good estimate to the
number of solutions for CSP benchmarks of model RB. Even if the
instances scales are relatively large, the estimates are found to be
over 63.129\% correct except the instance frb20-11-3. Therefore,
this experiment shows that PFCount is quite competitive compared
with the other counters.

\subsubsection{Evaluation the Performance of PFCount on Hard Instances}

\begin{table}
\centering
\includegraphics[scale=0.69]{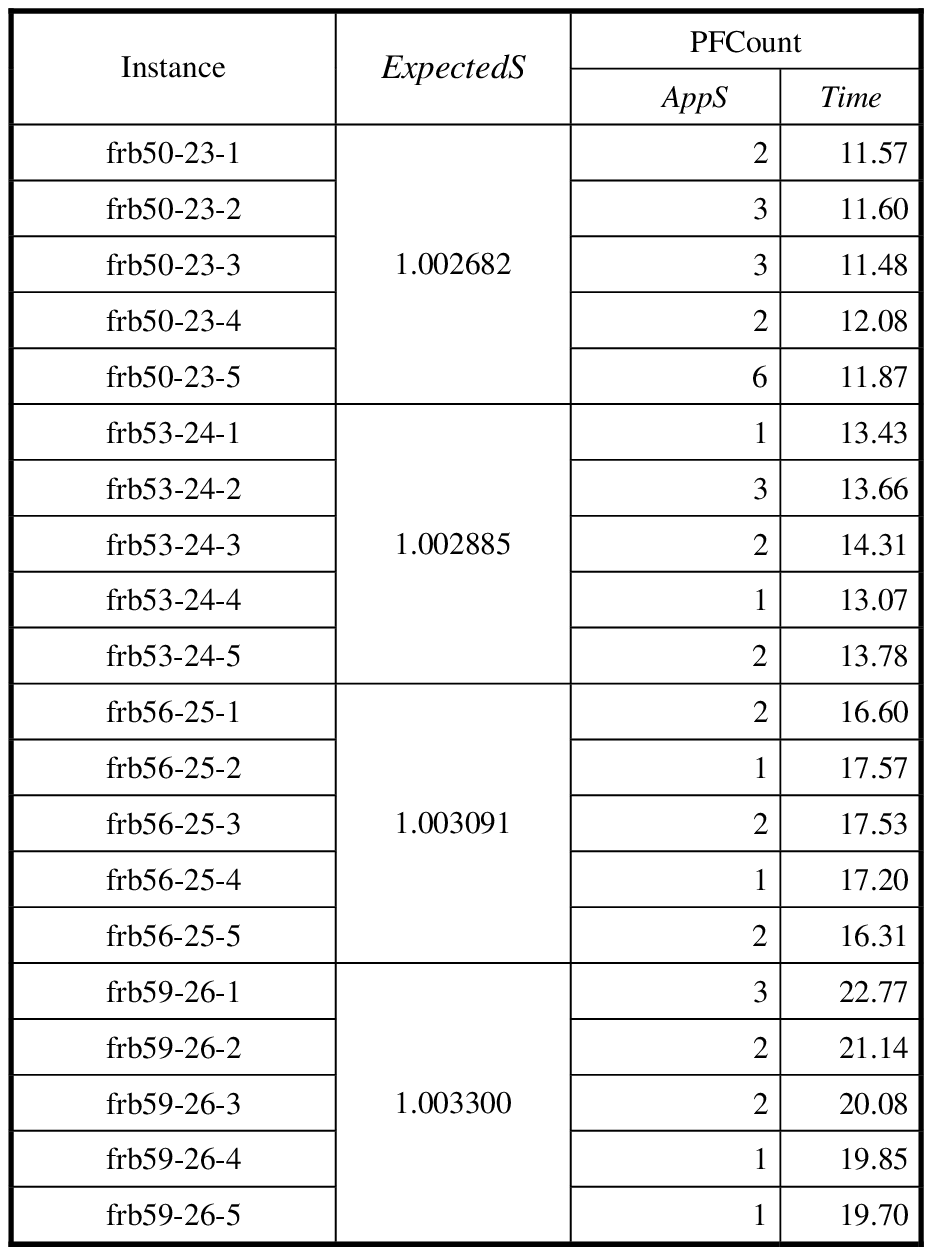}
\caption{Evaluation the performance of PFCount on hard CSP instances
(all times are in seconds).} \label{fig.example}
\end{table}

Table 2 presents the performance of our counter PFCount on hard
instances based on model RB\footnote{You can download at
http://www.nlsde.buaa.edu.cn/~kexu/benchmarks\\/benchmarks.htm.}.
These instances provide a challenge for experimental evaluation of
the CSP solvers. In the 1st international CSP solver competition in
2005, these instances can't be solved by all the participating CSP
solvers in 10 minutes. In the recent CSP solver competition, only
one solver can solve the instances frb50-23-1, frb50-23-4,
frb50-23-5, frb53-24-5, frb56-25-5, and frb59-26-5; only two solvers
can solve the instance frb53-24-3; four solvers can solve the
instance frb53-24-1; and the rest of instances still can't be solved
in 10 minutes. In Table 2, \textit {AppS} represents the approximate
number of solutions of each instance; \textit {ExpectedS} represents
the expected number of solutions of each instance, which can be
calculated as the following according to the definition of model RB
[19].

\begin{equation}
  ExpectedS=n^{\alpha n}(1-p)^{rn\ln n}
\end{equation}where \textit {n} denotes the number of variables;
$\alpha$ determines the domain size $d = n^\alpha$ of each variable;
\textit {r} determines the number $m = rn\ln n$ of constraints;
\textit {p} determines the number $t = pd^k$ of disallowed tuples of
each relation. When \textit {n} tends to infinite, \textit
{ExpectedS} is the number of solutions of the instances based on
model RB. Empirically, when \textit {n} is not very large, \textit
{ExpectedS} and the exact number of solutions are in the same order
of magnitude. Therefore, \textit {ExpectedS} precisely estimates the
number of solutions of the instances based on model RB. By analyzing
the results in Table 2, we can see that PFCount efficiently
estimates the number of solutions of these hard CSP instances. It
should be pointed out that our PFCount is only capable of estimating
the numbers of solutions rather enumerating the solutions.

\subsection{Evaluation on the Structural Instances}

In this subsection, we carry out experiments on graph coloring
instances from the DIMACS benchmark set\footnote{You can download at
http://mat.gsia.cmu.edu/COLOR02/.}. The \#CSP solvers compared with
PFCount are ILOG Solver 6.3 [20] and CSP+XORs. Table 3 illustrates
the results of the comparison of the \#CSP solvers on graph coloring
instances. In this table, \textit {AppS} is the approximate number
of solutions of each instance; \textit {ExactS} is the exact number
of solutions of each instance calculated by the exact counters. Note
that the results presented by the ILOG Solver and CSP+XORs are based
on [16] for lack of the binary codes. As can be seen from Table 3,
PFCount doesn't give good estimate on these structural instances in
contrast with the random instances. However, the run-time of PFCount
clearly outperforms other \#CSP solvers greatly.

\begin{table}
\centering
\includegraphics[scale=0.90]{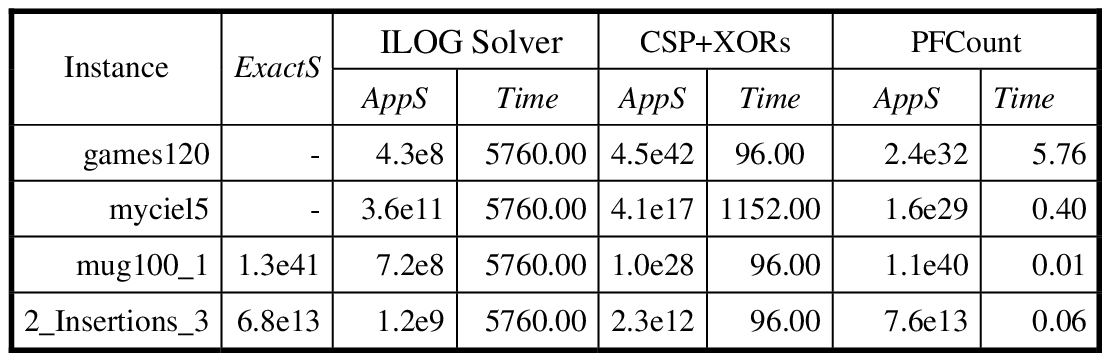}
\caption{Comparison of \#CSP solvers on graph coloring instances (-
represents out of time and times are in seconds).}
\label{fig.example}
\end{table}

\section{Conclusion}

This paper addresses a new approximate method for counting the
number of solutions for constraint satisfaction problem. It first
obtains the marginal probabilities of each variable and constraint
by the belief propagation approach, and then computes the number of
the solutions of a given CSP formula according to a partition
function, which obtained by introducing the free energy and
capturing the relationship between the probabilities of the
variables and the constraints. The experimental results also show
that the effectiveness of our method is much more evident especially
for larger instances close to the phase transition point.


\begin{thebibliography}{20}

\bibitem{proceeding1}Bayardo, R. J., and Pehoushek, J. D.: Counting models using
connected components. In: 17th National Conference on Artificial
Intelligence, pp. 157--162. AAAI Press, U.S. (2000)

\bibitem{proceeding2}Sang, T., Bacchus, F., Beame, P., Kautz, H. A., and Pitassi, T.: Combining component caching and clause learning for effective model counting.In: 7th Theory and Applications of Satisfiability Testing. Springer, Heidelberg (2004)

\bibitem{proceeding3}Darwiche, A.: New advances in compiling CNF into
decomposable negation normal form. In: 16th European Conference on
Artificial Intelligence, pp. 328--332. IOS Press, Washington (2004)

\bibitem{proceeding4}Thurley, M.: sharpSAT - counting models with advanced component
caching and implicit BCP. In: 9th Theory and Applications of
Satisfiability Testing, pp. 424--429. Springer, Heidelberg (2006)

\bibitem{proceeding5}Davies, J., and Bacchus, F.: Using more reasoning to improve \#SAT
solving. In: 22nd National Conference on Artificial Intelligence,
pp. 185--190. AAAI Press, U.S. (2007)

\bibitem{proceeding6}Wei, W., and Selman, B.: A new approach to model counting.
In: 8th Theory and Applications of Satisfiability Testing, pp.
324--339. Springer, Heidelberg (2005)

\bibitem{proceeding7}Gomes, C. P., Hoffmann, J., Sabharwal, A., and Selman, B.: From sampling to model counting. In: 20th International Joint Conference on Artificial Intelligence, pp. 2293--2299. Springer, Heidelberg (2007)

\bibitem{proceeding8}Gomes, C. P., Sabharwal, A., and Selman, B.: Model counting: A new
strategy for obtaining good bounds. In: 21st National Conference on
Artificial Intelligence, pp. 54--61. AAAI Press, U.S. (2006)

\bibitem{proceeding9}Gogate, V., and Dechter, R.: Approximate counting by sampling the
backtrackfree search space. In: 22nd National Conference on
Artificial Intelligence, pp. 198--203. AAAI Press, U.S. (2007)

\bibitem{proceeding10}Kroc, L., Sabharwal, A., and Selman, B.: Leveraging belief propagation, backtrack search, and statistics for model counting. In: 5th
Integration of AI and OR Techniques in Contraint Programming for
Combinatorial Optimzation Problems, pp. 127--141. Springer,
Heidelberg (2008)

\bibitem{proceeding11}Angelsmark, O., Jonsson, P., Linusson, S., and Thapper J.:
Determining the number of solutions to binary CSP instances. In: 8th
Principles and Practice of Constraint Programming, pp. 327--340.
Springer, Heidelberg (2002)

\bibitem{proceeding12}Bulatov, A., Dalmau, V.: Towards a dichotomy theorem for
the counting constraint satisfaction problem. In: 44th Annual IEEE
Symposium on Foundations of Computer Science, pp. 562--571. IEEE
Computer Society, Los Alamitos, CA (2003)

\bibitem{proceeding13}Pesant, G.: Counting solutions of CSPs: a structural
approach. In: International Joint Conference on Artificial
Intelligence, pp. 260--266. Springer, Heidelberg (2005)

\bibitem{jour14}Dyera, M., Goldbergb, L. A., Jerrum, M.: An
approximation trichotomy for Boolean \#CSP. J. Comput. Syst. Sci.
76(3-4), 267--277 (2010)

\bibitem{jour15}Yamakami, T.: Approximate counting for complex-weighted Boolean
constraint satisfaction problems. Inf. Comput. 219, 17--38 (2012)

\bibitem{proceeding16}Gomes, C. P., Hoeve, W. J., Sabharwal, A., and Selman, B.:
Counting CSP Solutions Using Generalized XOR Constraints. In:
National Conference on Artificial Intelligence, pp. 204--209. AAAI
Press, U.S. (2007)

\bibitem{proceeding17}Favier, A., Givry, S., Jegou, P.: Exploiting problem
structure for solution Counting. In: Principles and Practice of
Constraint Programming, pp. 335--343. Springer, Heidelberg (2009)

\bibitem{jour18}Zhang, P., Ramezanpour, A., Zdeborova, L., and Zecchina,
R.: Message passing for quantified Boolean formulas. J. Stat.
Mech-Theory E. pp. 05025 (2012)

\bibitem{jour19}Xu, K., Boussemart, F., Hemery, F., and Lecoutre C.: Random
constraint satisfaction: Easy generation of hard (satisfiable)
instances. Artif. Intell. 171(8-9), 514--534 (2007)

\bibitem{20}ILOG, SA. 2006. ILOG Solver6.3 reference manual.


\end{thebibliography}
\end{document}